\documentclass[10pt]{article}

\usepackage[margin=1in]{geometry}
\usepackage{microtype}
\usepackage{graphicx}
\usepackage{booktabs}
\usepackage{amsmath}
\usepackage{amssymb}
\usepackage{mathtools}
\usepackage{algorithm}
\usepackage{algorithmic}
\usepackage{placeins}
\usepackage{float}
\usepackage{caption}
\usepackage{authblk}
\usepackage[round]{natbib}
\usepackage[hidelinks]{hyperref}

\newcommand{\method}{MAP-Law}

\title{MAP-Law: Coverage-Driven Retrieval Control for Multi-Turn Legal Consultation}

\author[1]{Qinchuan Cheng}
\author[2]{Jiaqi Liu}
\author[3]{Ruixuan Xie}
\author[4]{Xiaoya Yuan}
\author[3]{Yuxin Liu}

\affil[1]{Xi'an Jiaotong University}
\affil[2]{Sichuan University}
\affil[3]{Southwestern University of Finance and Economics}
\affil[4]{Northeastern University}

\date{}

\begin{document}

\makeatletter
\twocolumn[
\begin{@twocolumnfalse}
\maketitle
\begin{abstract}
Legal consultation is inherently iterative: before giving advice, a system must identify relevant legal elements, gather missing facts and authorities, and determine whether the current evidence is sufficient. Existing retrieval-augmented legal agents often use fixed retrieval budgets or single-shot search, making them insensitive to the evolving coverage state of a consultation. This paper introduces \method, a coverage-driven retrieval-control framework for multi-turn legal consultation. \method\ maintains a structured map over user facts, legal elements, retrieval goals, and retrieved evidence, and uses element coverage, evidence validity coverage, and marginal retrieval gain to decide whether to retrieve, clarify, reformulate, or stop. On a 50-case synthetic Chinese labor-law consultation pilot with fixed legal-element schemas, a DeepSeek V4-Pro action-selection variant achieves full measured element coverage under the pilot metric while requiring 3.4 retrieval rounds and 7.1 evidence snippets on average. Diagnostic analyses show that model-backed action selection recovers rule-policy failure cases with a small retrieval-budget increase, while forced continuation mainly increases token and latency costs. These results suggest that legal-element coverage is a useful control signal for adaptive legal retrieval, while remaining bounded to retrieval-control behavior under synthetic fixed-schema conditions rather than deployment-level legal correctness.
\end{abstract}

\vspace{0.3in}
\end{@twocolumnfalse}
]
\makeatother

\section{Introduction}

Legal consultation is not a single-step question-answering problem. In a real consultation, a lawyer rarely provides advice immediately after reading the first user query. Instead, the lawyer identifies the relevant legal issue, decomposes it into legally required elements, asks follow-up questions when facts are missing, retrieves statutes or cases when authority is needed, and only then gives advice. This process is inherently iterative and coverage-driven: the key question is not only whether some relevant documents have been retrieved, but whether the legally required elements have been sufficiently supported by facts and evidence.

This observation creates an important challenge for AI-for-law systems. Retrieval-augmented generation (RAG) has become a common way to ground legal language models in statutes, cases, and domain corpora. However, many retrieval-augmented legal agents still rely on single-shot search or fixed retrieval budgets. Such systems retrieve a predetermined amount of evidence regardless of whether the consultation state is already sufficient or still missing a decisive legal element. As a result, they may stop too early and produce doctrinally incomplete answers, or retrieve too much and introduce cost, latency, and irrelevant context. In legal settings, both failure modes are problematic: missing one required element may change the legal conclusion, while excessive retrieval can obscure the basis of the final recommendation.

The broader AI4Law motivation is therefore not only to make legal language models more fluent, but to make their reasoning process more structured, controllable, and auditable. Legal consultation requires a bridge between legal task structure and language-agent behavior. A useful legal agent should know which legal elements remain unresolved, which facts and authorities have already been collected, which retrieval goal should be pursued next, and when additional retrieval is unlikely to improve the consultation state. This paper studies this control problem: can legal-element coverage serve as an explicit signal for adaptive retrieval and stopping in multi-turn legal consultation?

We present \method, a coverage-driven retrieval-control framework for multi-turn legal consultation. \method\ maintains a structured map over user facts, legal elements, retrieval goals, and retrieved evidence. At each round, the system estimates element coverage, evidence validity coverage, and marginal retrieval gain, then selects the next action: retrieving statutes, retrieving cases, requesting clarification, reformulating the query, or stopping. In this design, retrieval is not treated as a fixed-depth preprocessing step. Instead, it becomes an adaptive process guided by the legal elements that remain unsupported.

Our study focuses on a fixed-schema synthetic Chinese labor-law consultation pilot. This setting is intentionally controlled: legal-element initialization is held fixed so that the experiments isolate the downstream retrieval-control problem. Given the same validated set of legal elements, we ask whether different controllers can retrieve evidence selectively, recover missing elements, and stop before unnecessary retrieval accumulates. This does not solve end-to-end legal consultation, but it targets a key subproblem for reliable legal agents: controlling the evidence-collection process once the legal task structure is available.

We make three contributions. First, we formulate multi-turn legal consultation as adaptive evidence collection over legal-element coverage, providing a structured interface between legal reasoning and language-agent control. Second, we propose \method, which couples a legal-element plan graph, an evidence graph, a coverage monitor, and an action selector to decide when to retrieve, clarify, reformulate, or stop. Third, we evaluate \method\ on a 50-case synthetic Chinese labor-law pilot and conduct diagnostic analyses including case-type failure analysis, stopping-threshold sensitivity, hard-case budget sweeps, and multi-seed robustness checks. The results show that model-backed coverage control achieves full measured element coverage under the fixed-schema pilot metric while using fewer retrieval rounds and evidence snippets than fixed-round or deterministic alternatives.

The conclusions are deliberately bounded but practically meaningful. The reported coverage scores are internal retrieval-control metrics under synthetic fixed-schema conditions; they should not be interpreted as deployment-level legal correctness. Nevertheless, the study highlights a general design principle for AI4Law: legal agents should not only retrieve relevant text, but also monitor whether legally required elements have been covered. By making the consultation state explicit, \method\ provides a controllable and auditable interface for multi-turn legal retrieval, and offers a step toward legal AI systems that are more structured, efficient, and trustworthy.

\begin{figure*}[t]
\centering
\includegraphics[width=0.88\textwidth]{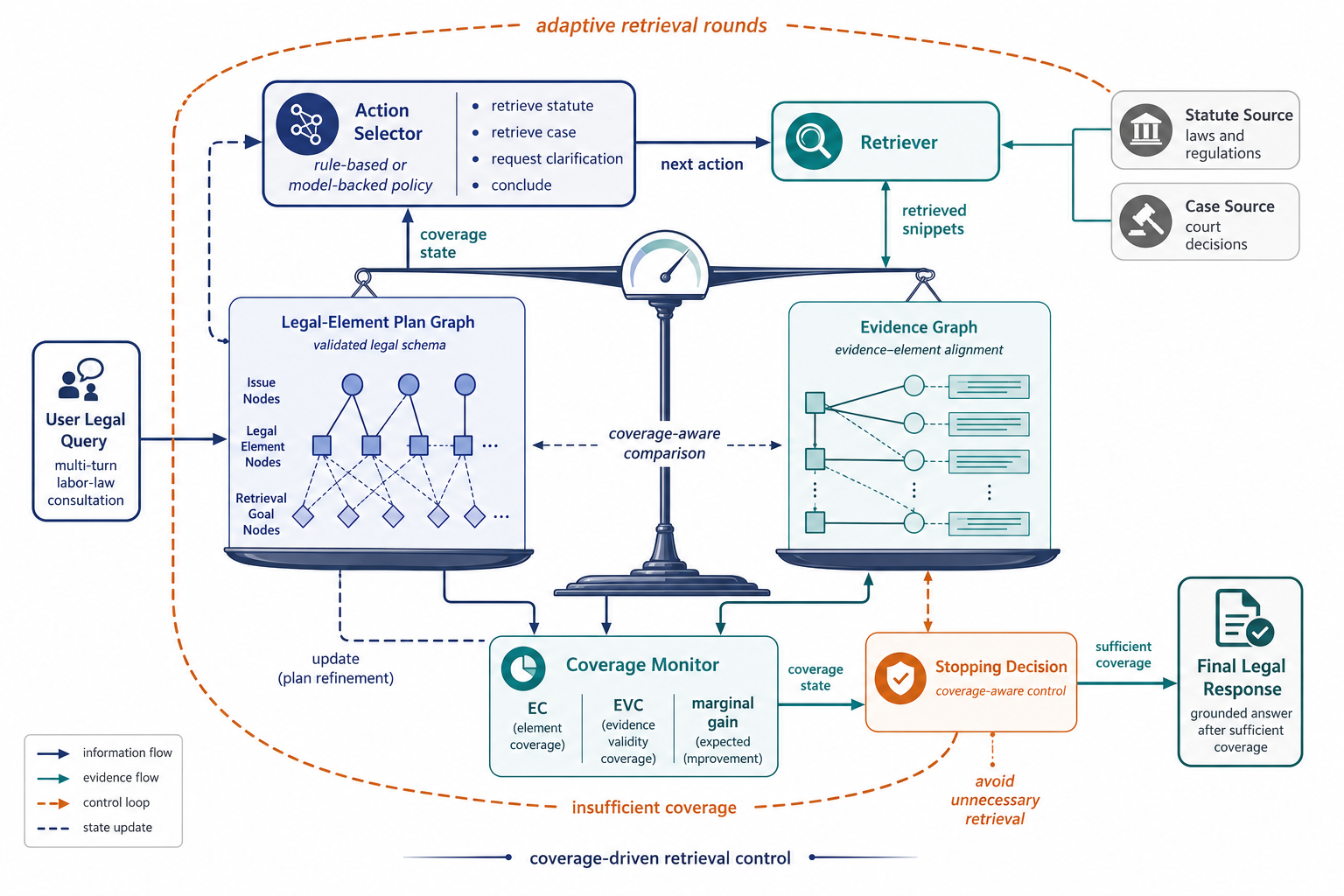}
\caption{Overview of \method. The system maintains a legal-element plan graph and an evidence graph, monitors coverage, selects the next retrieval or clarification action, and stops once the consultation state is sufficiently covered.}
\label{fig:framework}
\end{figure*}

\section{Related Work}

\paragraph{Legal language models and legal benchmarks.}
Legal NLP has moved from document classification and entailment toward broader
legal reasoning, retrieval, and evaluation benchmarks. Early and widely used
resources include LegalBERT and LexGLUE for legal language understanding
\citep{chalkidis2020legalbert,chalkidis2022lexglue}, JEC-QA for legal-domain
question answering \citep{zhong2020jecqa,zhong2020jecqa_dataset}, and more
recent benchmark suites such as LegalBench, LawBench, and LegalBench-RAG
\citep{guha2023legalbench,fei2023lawbench,legalbenchrag}. Related datasets
also cover case retrieval, holdings, contract review, contract entailment, and
legal information extraction/entailment
\citep{ma2021lecard,zheng2021casehold,hendrycks2021cuad,koreeda2021contractnli,rabelo2022coliee}.
Recent corpus-building efforts further expand the scale and multilingual
coverage of legal text resources \citep{henderson2022pileoflaw,niklaus2024multilegalpile}.
These resources are valuable, but many focus on single-shot prediction,
retrieval, or answer evaluation. Multi-turn consultation adds a control
problem: deciding which legally relevant facts or authorities are still missing
before the system should answer.

\paragraph{Legal RAG and agentic consultation systems.}
Retrieval-augmented generation and dense retrieval ground language generation
in external evidence \citep{karpukhin2020dpr,lewis2020rag}. Legal RAG systems
extend this idea to statutes, cases, and domain-specific legal corpora
\citep{chatlaw2023,lexrag}. Meanwhile, agentic language-model methods show
that models can interleave reasoning, search, tool use, and self-reflection
\citep{wei2022cot,press2022selfask,yao2023react,yao2023tree,schick2023toolformer,shinn2023reflexion}.
In law, however, tool use and retrieval should be constrained by legal elements
and evidence quality. A fluent answer can still be doctrinally incomplete or
factually unsupported if the system has not checked the necessary legal
conditions.

\paragraph{Evaluation and control in high-stakes legal settings.}
Evaluation for law must consider doctrinal correctness, factual alignment,
robustness, user impact, and the risk of over-trust \citep{surden2019ai}. This
paper studies a narrower control problem inside that broader agenda: whether a
legal consultation agent can use legal-element coverage as an explicit signal
for retrieval, clarification, and stopping. Rather than treating retrieval as a
fixed-depth preprocessing step, \method\ treats evidence collection as an
adaptive process over a structured legal-element state.

\section{Method}

\method\ represents a consultation state as a directed map
\[
G_t = (F_t, E, R_t, A_t),
\]
where $F_t$ is the set of user-provided facts after turn $t$, $E$ is a set of legal elements for the case type, $R_t$ is retrieved legal evidence, and $A_t$ links facts and evidence to elements. The agent repeats three steps.

\begin{table}[t]
\centering
\caption{Core notation used in MAP-Law.}
\label{tab:notation}
\small
\begin{tabular}{ll}
\toprule
Symbol & Meaning \\
\midrule
$F_t$ & user facts observed by round $t$ \\
$E$ & legal-element set for the case type \\
$R_t$ & retrieved evidence by round $t$ \\
$A_t$ & fact/evidence-to-element links \\
$\mathrm{EC}_t$ & element coverage at round $t$ \\
$\mathrm{EVC}_t$ & evidence validity coverage at round $t$ \\
$\mathrm{MG}_t$ & marginal gain at round $t$ \\
$\theta_e,\theta_{ev},\delta$ & stopping thresholds \\
$r_{\max}$ & maximum retrieval budget \\
\bottomrule
\end{tabular}
\end{table}

\paragraph{Graph semantics.}
The map has four node families. Issue nodes represent the high-level legal
question under discussion. Element nodes represent legally required subclaims,
such as whether an employment relationship exists or whether a dismissal ground
is lawful. Retrieval-goal nodes specify what evidence is still missing for a
particular element. Evidence nodes store retrieved statutes, cases, extracted
facts, or web-style policy notes. Element nodes carry both a support status
(\texttt{unsupported}, \texttt{partially\_supported},
\texttt{fully\_supported}) and short missing-evidence descriptions, so the map
can be used not only for post-hoc explanation but also for action selection.

\paragraph{Element-aware query planning.}
At each round, the agent identifies unsupported or partially supported elements
and selects a target element together with an action from the space
\{\texttt{retrieve\_statute}, \texttt{retrieve\_case},
\texttt{request\_clarification}, \texttt{generate\_conclusion}\}. A
model-backed policy rewrites the query from the current dialogue and map state;
the deterministic rule policy falls back to element labels plus missing-evidence
descriptions. This design makes the retrieval target explicit: the query is not
generated from the full dialogue alone, but from the subset of legal elements
that remain unresolved.

\paragraph{Evidence alignment.}
Retrieved evidence is linked to legal elements through lightweight lexical and semantic matching. For each element $e \in E$, the system marks it as covered when the current facts and retrieved evidence jointly satisfy the element criterion. This produces an element-coverage score:
\[
\mathrm{EC}(G_t) = \frac{1}{|E|} \sum_{e \in E} \mathbb{I}[e\ \mathrm{covered}].
\]
We also track evidence validity coverage (EVC), the proportion of selected evidence that can be matched to at least one legal element.

\paragraph{State update and audit trail.}
Every retrieval action appends evidence nodes to the graph, records the
retrieval round, and links evidence to one or more target elements when the
matching score is sufficient. This produces an explicit audit trail:
the system can report which legal elements were considered unresolved at each
round, which query was issued in response, which authorities were retrieved,
and why the final stopping condition was triggered. In a legal setting, this
trace is useful both for analysis and for downstream human review, because the
control decision can be reconstructed from visible state rather than hidden
inside a single model output.

\paragraph{Adaptive stopping.}
The stopping rule combines sufficiency and marginal gain. Let $\mathrm{MG}_t$
denote the average overlap-based gain of the evidence retrieved at round $t$
with respect to the currently unresolved elements. The agent stops if
\[
(\mathrm{EC}_t \ge \theta_e \wedge \mathrm{EVC}_t \ge \theta_{ev})
\;\;\vee\;\;
(\mathrm{MG}_{t-1} < \delta \wedge \mathrm{MG}_t < \delta)
\]
or when the maximum round budget is exhausted. In the reported runs,
$\theta_e=0.85$, $\theta_{ev}=0.70$, and $\delta=0.05$. This differs from
fixed-round RAG, which performs the same number of retrieval rounds for every
case. The design goal is to spend turns only where the legal map indicates
missing elements and to terminate once additional retrieval has become legally
unproductive.

\paragraph{Controller decomposition.}
MAP-Law is designed so that its main components can be separated analytically.
The legal map determines which issues and legal elements remain unresolved. The
action selector determines where the next retrieval step should go. The
stopping rule determines when the current evidence state is sufficient to stop.
This decomposition is reflected in the ablations: MAP-no-graph removes the
structured state, MAP-no-threshold-stop removes the explicit stopping check,
and the rule policy replaces model-backed action selection with deterministic
heuristics. The empirical study therefore probes the contribution of each part
of the controller rather than treating MAP-Law as an indivisible black box.

\begin{algorithm}[t]
\caption{\method\ retrieval-control loop}
\label{alg:maplaw}
\begin{algorithmic}[1]
\STATE Initialize issue nodes, legal-element nodes, and retrieval goals
\STATE Set round counter $t \leftarrow 0$
\WHILE{$t < r_{\max}$}
    \STATE Compute $\mathrm{EC}_t$ and $\mathrm{EVC}_t$ from the current graph
    \STATE Compute recent marginal gains over unresolved elements
    \IF{$(\mathrm{EC}_t \ge \theta_e \wedge \mathrm{EVC}_t \ge \theta_{ev})$ or recent marginal gains are $< \delta$}
        \STATE \textbf{break}
    \ENDIF
    \STATE Select target element and next action from the MAP state
    \IF{action is \texttt{generate\_conclusion}}
        \STATE \textbf{break}
    \ENDIF
    \STATE Retrieve statutes/cases or request clarification
    \STATE Link new evidence to candidate elements and update support status
    \STATE $t \leftarrow t + 1$
\ENDWHILE
\STATE Generate final response from the resulting consultation state
\end{algorithmic}
\end{algorithm}

\section{Experimental Setup}

\begin{figure}[t]
\centering
\includegraphics[width=0.48\textwidth]{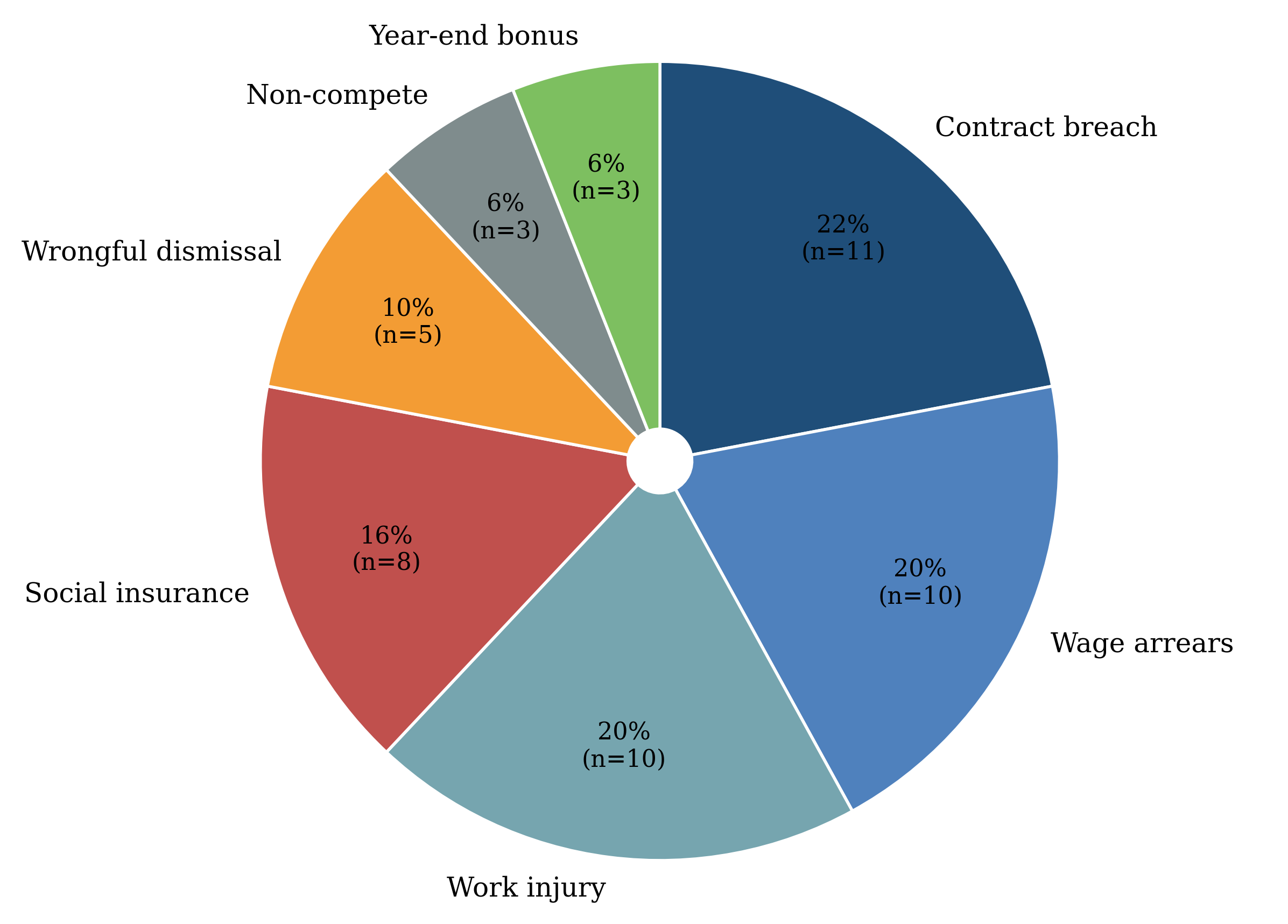}
\caption{Case-type composition of the 50-case synthetic labor-law pilot. The pilot is intentionally imbalanced toward common labor-law dispute categories so that both routine and hard categories appear in the diagnostic analysis.}
\label{fig:dataset-composition}
\end{figure}

\paragraph{Benchmark.}
We use a 50-case synthetic Chinese labor-law consultation pilot constructed around common dispute types: wage arrears, work injury, wrongful dismissal, social insurance, contract breach, non-compete, and year-end bonus. Each case includes a short user query, a case type, and legal elements expected to be covered during consultation.
The cases are synthetic and are used to test retrieval-control behavior rather than to evaluate final legal advice quality. Figure~\ref{fig:dataset-composition} summarizes the case-type composition of this pilot.

\paragraph{Systems.}
We compare seven systems. \textbf{\method\ (V4-Pro action)} uses DeepSeek V4-Pro for action selection with fixed legal-element initialization. \textbf{\method\ (Rule)} uses the same map and metrics but a deterministic fallback policy. \textbf{MAP-no-threshold-stop} removes the explicit coverage and marginal-gain stopping check while keeping the same action policy, including its ability to emit a conclusion action. \textbf{MAP-no-graph} removes the legal-element map. \textbf{Fixed-3}, \textbf{Fixed-5}, and \textbf{Fixed-7} use fixed retrieval budgets.

\paragraph{Retriever and evidence inventory.}
All systems share the same simulated Chinese legal corpus. Statute retrieval
returns up to three items per action, case retrieval returns up to two items,
and evidence can be linked to one or more legal elements. This controlled setup
ensures that the main comparison isolates retrieval control rather than changes
in the underlying corpus.

\paragraph{Why fixed legal-element initialization.}
The paper intentionally fixes legal-element initialization across all reported
systems. This design removes one major source of variance and forces the
comparison to focus on retrieval control: given the same consultation goals,
which controller retrieves the right evidence with the lowest cost? The fixed
schema setting is therefore not a simplification hidden in the appendix; it is
part of the experimental claim. The paper argues that MAP-Law is already useful
when legal elements are validated or supplied by an upstream legal-analysis
module, even before fully automatic legal planning is solved.

\paragraph{Metrics.}
The main metric is element coverage (EC), which measures whether legally required elements are covered. EVC measures whether selected evidence is legally connected to elements. We also report average rounds and average evidence count as cost proxies.

\paragraph{Interpretation of graph-free baselines.}
EC in this implementation measures actionable element coverage maintained by
the controller. Graph-free fixed-retrieval baselines may retrieve topically
relevant evidence, but they do not maintain or update element-level support
status. Their EC therefore evaluates the availability of an explicit coverage
state rather than oracle post-hoc relevance of all retrieved documents. A
stronger future evaluation should add post-hoc legal-element alignment for
graph-free baselines to separate retrieval recall from controller-state
availability.

\paragraph{Reproducibility note.}
The rule, no-threshold-stop, no-graph, and fixed-round rows are offline deterministic runs. The V4-Pro row uses the same legal-element template initialization but calls DeepSeek V4-Pro for action selection. We keep legal-element initialization fixed so that all systems are evaluated against the same consultation goals.

\paragraph{Research questions.}
The experimental section is organized around three questions. \textbf{RQ1} asks
whether coverage-driven retrieval control improves the efficiency--completeness
balance relative to fixed-round retrieval. \textbf{RQ2} asks which components
matter most for that improvement: the legal map, the stopping rule, or the
action-selection policy. \textbf{RQ3} asks whether the strongest system wins by
retrieving more evidence overall, or by retrieving more selectively on legally
hard cases.

\paragraph{Additional diagnostics.}
To strengthen internal validity, we run a hard-case budget sweep on the 11 dialogues that fail under the original deterministic rule run. This extension compares the normal policy against a hard-no-stop variant that is forbidden from emitting \texttt{generate\_conclusion} before the round budget is exhausted. We test budgets $r\in\{1,2,3,5,7\}$ for both the rule policy and Qwen-Plus, which is used here because it matches the main fixed-schema coverage result while offering lower latency than the other additional models. To test robustness to dataset composition, we also run a multi-seed study on four independently generated 30-case synthetic pilots with seeds $\{42,7,21,84\}$.

\section{Results}

\begin{table*}[t]
\centering
\caption{Main synthetic-pilot results. Panel A compares \method\ against deterministic ablations and fixed-round baselines. Panel B swaps the action-selection model under the same fixed-schema MAP configuration. EC/EVC report measured coverage under this fixed-schema synthetic pilot and should not be interpreted as legal correctness. Lower rounds, evidence, tokens, and latency indicate lower interaction cost.}
\label{tab:main-results}
\small
\begin{tabular}{lcccccc}
\toprule
\multicolumn{7}{c}{Panel A: Main systems and ablations} \\
\midrule
System & EC & EVC & Rounds & Evidence & Tokens & Latency (s) \\
\midrule
\method\ (V4-Pro action) & 1.000 & 1.000 & 3.36 & 7.08 & 2565.64 & 20.33 \\
\method\ (Rule) & 0.780 & 1.000 & 4.66 & 11.64 & -- & -- \\
MAP-no-threshold-stop & 0.780 & 1.000 & 4.66 & 11.64 & -- & -- \\
MAP-no-graph & 0.000 & 0.600 & 3.00 & 6.00 & -- & -- \\
Fixed-3 & 0.000 & 1.000 & 3.00 & 8.00 & -- & -- \\
Fixed-5 & 0.000 & 1.000 & 5.00 & 13.00 & -- & -- \\
Fixed-7 & 0.000 & 1.000 & 7.00 & 18.00 & -- & -- \\
\midrule
\multicolumn{7}{c}{Panel B: Cross-model action selection under the same fixed schema} \\
\midrule
Action selector & EC & EVC & Rounds & Evidence & Tokens & Latency (s) \\
\midrule
DeepSeek V4-Pro & 1.000 & 1.000 & 3.36 & 7.08 & 2565.64 & 20.33 \\
Qwen-Plus & 1.000 & 1.000 & 3.36 & 7.08 & 3019.40 & 18.24 \\
GLM-5.1 & 1.000 & 1.000 & 3.36 & 7.08 & 3166.38 & 43.94 \\
MiniMax-M2.7 & 1.000 & 1.000 & 3.36 & 7.08 & 4136.36 & 59.85 \\
\bottomrule
\end{tabular}
\end{table*}

The main 50-case result is summarized in Table~\ref{tab:main-results}. The
V4-Pro action-selection variant reaches EC/EVC 1.000 while using 3.4 rounds and
7.1 evidence snippets on average. By contrast, the deterministic rule policy
and MAP-no-threshold-stop both remain at EC 0.780 with 4.7 rounds and 11.6
evidence snippets, while MAP-no-graph fails on EC altogether. Fixed-round
baselines retrieve more evidence as the budget grows, but they do not improve
EC in this implementation because they lack legal-element control. This
indicates that the legal map is not just an interpretability artifact; it is
part of the control signal.

This directly answers RQ1. In the current pilot, stronger behavior does not
come from deeper fixed retrieval. It comes from selectively targeting unresolved
legal elements and terminating once coverage has saturated. Relative to the
fixed seven-round baseline, the strongest MAP-Law configuration reduces average
evidence volume from 18.0 to 7.08 and average rounds from 7.0 to 3.36 while
also achieving the best measured coverage.

The offline rule-only run gives a sharper view of the remaining bottleneck. After fixing the Chinese marginal-gain estimator, rule MAP and MAP-no-threshold-stop both reach EC 0.780 with 4.7 rounds, which means removing the threshold stop alone does not solve the hard categories. The V4-Pro row suggests that targeted action selection, rather than simply increasing the number of turns, is what recovers the remaining coverage gap.

A same-seed two-run repeat check of the V4-Pro action-selection configuration
also preserves EC/EVC 1.000. Table~\ref{tab:main-results} additionally shows
that Qwen-Plus, GLM-5.1, and MiniMax-M2.7 all match the same EC/EVC and average
rounds as V4-Pro under fixed legal-element initialization. The remaining
differences are concentrated in token cost and latency. Under this pilot
metric, the control architecture appears to dominate once the action selector
is competent enough, while model choice primarily shifts efficiency.

This cross-model consistency strengthens the paper's central claim. MAP-Law is
not best understood as a result tied to one preferred proprietary model.
Instead, the fixed-schema experiments suggest that once the action selector is
competent enough, the dominant factor is the structured control architecture:
the model is asked to operate inside a legal-element state, rather than to
solve the consultation from raw dialogue alone.

\paragraph{Hard-case budget sweep.}
The strongest stopping result comes from the 11 hard cases that the original rule baseline fails completely. Here the deterministic rule policy remains at EC 0.000 for every tested budget from $r=1$ through $r=7$, so additional budget does not make those cases solvable. Qwen-Plus behaves very differently: EC rises from 0.545 at $r=1$ to 1.000 at $r=2$, and then remains at 1.000 for $r=3,5,7$. This is the behavior we would expect from a useful stopping mechanism: a small budget increase is enough to recover the missing legal elements, but once coverage has saturated, extra retrieval only raises cost. Under the normal Qwen-Plus policy, average token usage is 1311 at $r=2$ and about 2200 at $r=5$ and $r=7$ because the agent usually concludes early. Under hard-no-stop, by contrast, token usage rises to 5006 at $r=5$ and 8461 at $r=7$, with average latency increasing to 29.7s and 55.7s. The evidence count shows the same pattern. Stopping therefore should be interpreted as a cost-control mechanism whose value emerges after competent action selection has already made the hard cases recoverable.

\begin{table}[t]
\centering
\caption{Compact summary of the hard-case budget sweep on the 11 rule-failure cases. This is a diagnostic subset selected from failures of the original rule run rather than an unbiased average-case test. Detailed evidence and token costs are reported in Table~\ref{tab:appendix-hardcase} in the appendix.}
\label{tab:hardcase-compact}
\footnotesize
\setlength{\tabcolsep}{4pt}
\begin{tabular}{@{}lccc@{}}
\toprule
Policy & Budget & EC & Latency (s) \\
\midrule
Rule & 1 & 0.000 & -- \\
Rule & 7 & 0.000 & -- \\
Qwen-Plus & 1 & 0.545 & 5.20 \\
Qwen-Plus & 2 & 1.000 & 13.83 \\
Qwen-Plus & 7 & 1.000 & 17.77 \\
Qwen no-stop & 2 & 1.000 & 10.92 \\
Qwen no-stop & 7 & 1.000 & 55.74 \\
\bottomrule
\end{tabular}
\end{table}

\paragraph{Multi-seed robustness.}
The multi-seed study gives a stronger robustness check than the same-seed API repeats because the synthetic case mixture changes across runs. Here the rule baseline is visibly unstable: EC varies from 0.533 to 0.733 across the four 30-case pilots, with average rounds between 4.8 and 5.4. By contrast, the Qwen-Plus model-backed policy remains fully stable under the same evaluation setup, preserving EC/EVC 1.000 on all four seeds while keeping average rounds between 2.97 and 3.30. This does not establish real-world robustness, since all four pilots are still synthetic, but it does show that the current model-backed control pattern is substantially less sensitive to dataset composition than the deterministic baseline.

Together, the hard-case sweep and the multi-seed study answer RQ3 more sharply
than the main table alone. The best-performing controller is not simply
retrieving more evidence overall. On hard cases, it needs only a small increase
in budget to recover the missing legal elements, after which additional
retrieval is mostly wasteful. Across different synthetic case mixtures, the
same policy remains stable while the deterministic baseline varies
substantially. The qualitative picture is therefore one of retrieval
discipline, not brute-force retrieval depth.

\begin{figure}[t]
\centering
\includegraphics[width=0.48\textwidth]{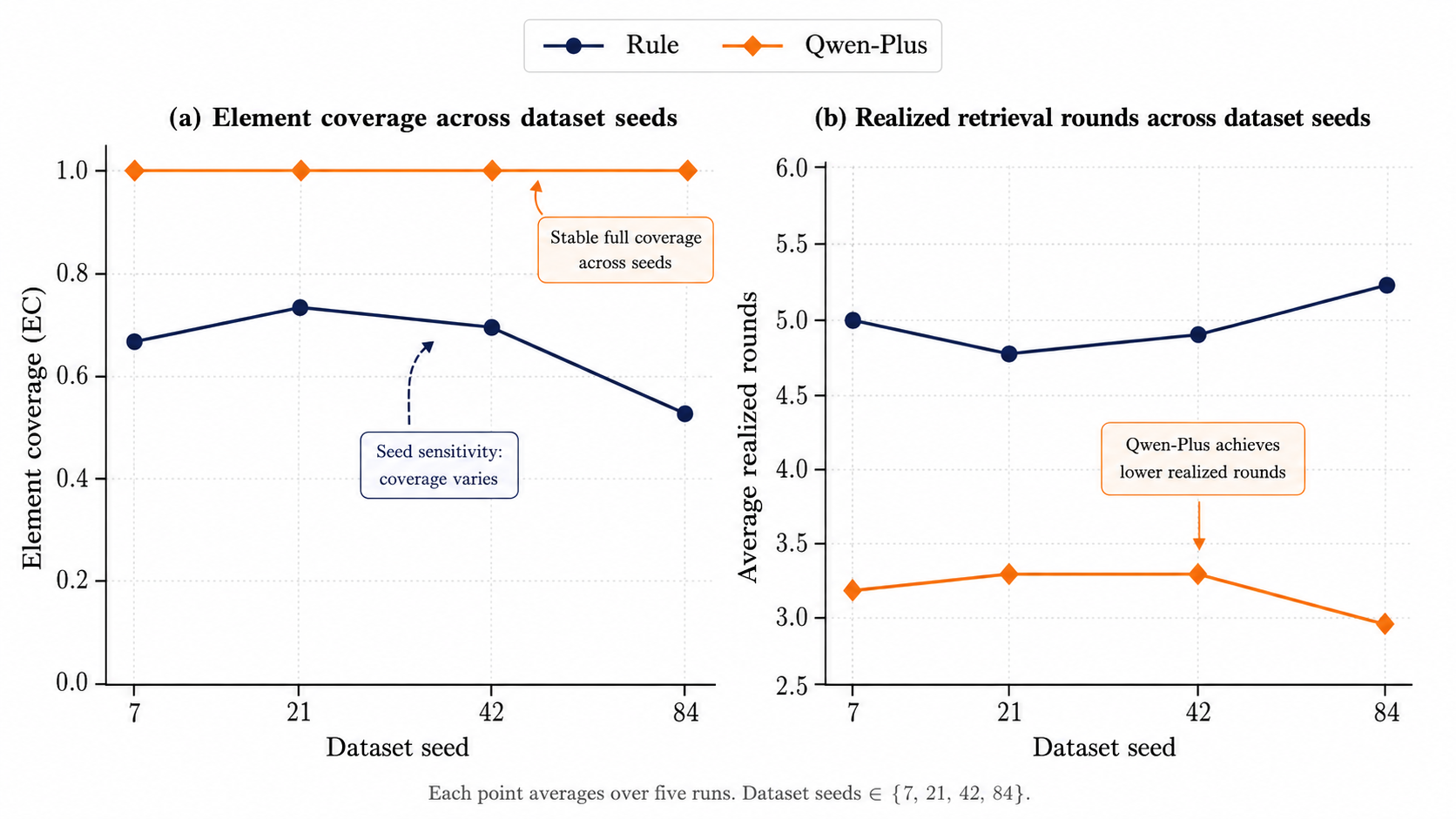}
\caption{Multi-seed robustness over independently generated 30-case synthetic pilots. This figure measures robustness to synthetic dataset composition, not real-world legal robustness. The rule baseline varies substantially across seeds, while Qwen-Plus remains stable under the same fixed-schema evaluation setting.}
\label{fig:seed-robustness-main}
\end{figure}

\section{Stopping and Budget Trade-Off}

\begin{figure}[t]
\centering
\includegraphics[width=0.48\textwidth]{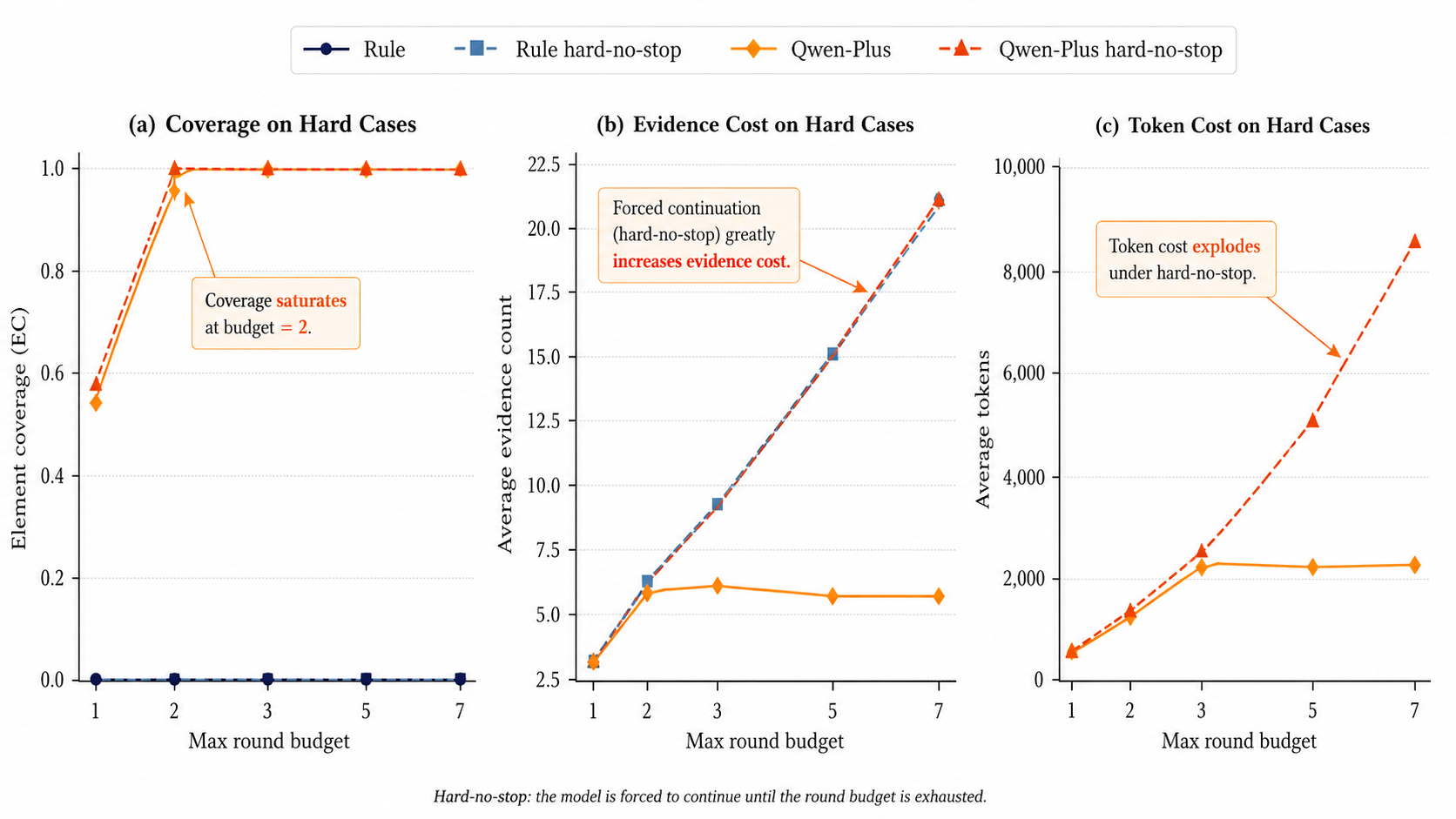}
\caption{Hard-case budget sweep on the 11 diagnostic rule-failure cases. The deterministic rule policy remains at zero measured element coverage across all budgets, while Qwen-Plus reaches full measured coverage by budget 2 and then mainly incurs additional cost when forced to continue retrieval.}
\label{fig:hardcase-budget}
\end{figure}

Figure~\ref{fig:hardcase-budget} visualizes the hard-case budget sweep reported
above. The key observation is that stopping is not what makes hard cases
solvable: the rule policy remains at zero coverage even as the budget increases.
Instead, once a model-backed selector recovers the missing legal elements,
adaptive stopping prevents unnecessary evidence and token growth. We therefore
interpret the stopping rule primarily as a cost-control mechanism that becomes
valuable after action selection is sufficiently element-aware.

For completeness, we also checked threshold sensitivity for the rule-only
policy. After replacing English word overlap with character-level overlap for
Chinese legal text, the threshold sweep over
$\delta\in\{0,0.01,0.03,0.05,0.08\}$ is stable: all settings reach EC 0.780
with 4.7 rounds and 11.6 evidence snippets on average. We therefore treat
$\delta$ as a calibration parameter rather than the main driver of the result;
the threshold figure is moved to the appendix.

\section{Case-Type Diagnostic Analysis}

We further group EC and average retrieval rounds by case type for
\method\ (Rule), MAP-no-threshold-stop, and V4-Pro action selection. The full
diagnostic matrix is reported in Figure~\ref{fig:appendix-case-diagnostics} in the appendix.
The diagnostic shows that the deterministic policies cover routine categories
such as wage arrears, work injury, wrongful dismissal, year-end bonus, and
contract-breach disputes, but fail on social-insurance and non-compete cases.
Because MAP-no-threshold-stop exhibits the same failures as rule MAP, these
errors are not primarily caused by premature threshold stopping. Instead, they
suggest that the rule query policy fails to retrieve evidence aligned with
more specialized legal elements.

By contrast, V4-Pro action selection recovers full measured coverage on these
hard categories while using fewer retrieval rounds under the same fixed
legal-element initialization. This supports the interpretation from the
hard-case budget sweep: the main bottleneck is not raw retrieval depth, but
element-aware action selection and evidence-element alignment. The result also
motivates future work on calibrated query rewriting and expert-validated
evidence linking for specialized legal categories.

\section{Discussion}

The main empirical lesson of MAP-Law is that, for legal consultation, better
performance does not primarily come from retrieving more material. It comes
from retrieving the \emph{right} material under an explicit model of what legal
support is still missing. This distinction matters because legal consultation
is not an ordinary search problem. A consultation may already contain many
topically relevant authorities while still missing one decisive legal element;
conversely, continuing to retrieve after the key elements are supported may add
cost, redundancy, and distraction without improving the answer.

The ablations and diagnostics make this point concrete. Fixed-round baselines
retrieve more evidence as the budget grows, but they do not recover actionable
element coverage without a legal-element state. The no-threshold-stop variant
matches the rule policy, suggesting that the observed rule failures are not
mainly caused by the explicit threshold condition. The hard-case budget sweep
further shows that increasing the budget helps only until the model-backed
selector has recovered the missing legal elements; after that point, more
retrieval mostly increases token cost and latency. The case-type diagnostic in Figure~\ref{fig:appendix-case-diagnostics} in the appendix is consistent with this view:
deterministic control handles routine categories but fails on social-insurance
and non-compete disputes, where more element-aware action selection is needed.

This perspective also helps explain why the cross-model results are so
consistent. Once the legal control interface is strong enough, several
different frontier models behave similarly on the pilot metric. That does not
mean model choice is irrelevant, but it does suggest that process design is a
first-order variable in legal AI. For workshop purposes, this is the broader
takeaway of the paper: improvements in legal agents may come as much from
better state representation and stopping discipline as from scaling the
underlying language model.

\section{Limitations}

This study is a fixed-schema synthetic pilot rather than a deployment
evaluation. The 50-case benchmark is small, synthetic, and limited to Chinese
labor-law consultation, so the results should be interpreted as evidence about
retrieval-control behavior under controlled conditions, not as a general claim
about legal consultation quality.

The experiments fix legal-element initialization to isolate the downstream
control problem: given validated legal elements, can the agent retrieve
selectively and stop appropriately? In real applications, such schemas must be
provided or verified by expert templates, jurisdiction-specific rules, or
upstream issue-spotting modules.

The current evidence matcher is lightweight, so EC may over-credit or
under-credit evidence-element alignment in realistic legal texts. Future work
should use expert-calibrated legal entailment, citation verification, and human
legal review. Finally, the graph-free baselines cannot maintain actionable
element-level coverage, so future evaluations should add post-hoc
legal-element alignment for fixed-retrieval baselines.

\section{Conclusion}

\method\ provides preliminary evidence that legal-element coverage can serve as a practical control signal for multi-turn legal retrieval. The strongest model-backed variant improves measured coverage while keeping interaction cost low, and the expanded diagnostics sharpen the interpretation of that result. In particular, the hard-case budget sweep shows that additional budget helps only until the competent policy has covered the missing legal elements, after which stopping becomes mainly a cost-saving device; the multi-seed study further shows that model-backed control is materially more stable than the deterministic baseline across synthetic pilot draws. The main lesson is that adaptive retrieval control is useful, but its reliability depends on the quality of legal-element modeling, evidence alignment, and stopping calibration.

\clearpage
\onecolumn
\appendix

\section{Prompt and Experimental Configuration}

\subsection{Action Space and Prompt Interface}

All reported model-backed runs use the same action space:
\texttt{retrieve\_statute}, \texttt{retrieve\_case}, \texttt{search\_web},
\texttt{extract\_fact}, \texttt{request\_clarification}, and
\texttt{generate\_conclusion}. The action selector receives a serialized MAP
state containing the current user query, the round index, EC/EVC, issue nodes,
element nodes, retrieved evidence summaries, and pending retrieval goals. It
must return a JSON object with four fields:
\texttt{action}, \texttt{reasoning}, \texttt{target\_element}, and
\texttt{query}.

The core system prompt used for action selection is:

\begin{quote}\small
\texttt{You are a legal consultation planning expert. Given the current MAP}\\
\texttt{state (plan graph + evidence graph), select the next action. You must}\\
\texttt{output a JSON object with your reasoning and action choice.}\\
\texttt{Prioritize unsupported or partially supported elements. Follow the}\\
\texttt{plan graph and address open issues first. If missing-evidence}\\
\texttt{descriptions exist for an element, prioritize retrieving that evidence.}\\
\texttt{Only emit generate\_conclusion when element coverage is sufficient}\\
\texttt{(EC >= 0.7). Use request\_clarification only when the query is too}\\
\texttt{ambiguous to proceed.}
\end{quote}

The state prompt then enumerates: (i) issues with current support status,
(ii) legal elements with support counts and missing-evidence descriptions,
(iii) retrieved evidence with source type and relevance score, and
(iv) pending retrieval goals. This prompt format is shared by DeepSeek
V4-Pro, Qwen-Plus, GLM-5.1, and MiniMax-M2.7 in the fixed-schema setting.

\subsection{Plan Initialization Prompt}

The main paper reports only fixed legal-element initialization, because this
setting isolates retrieval control from schema-generation noise. For
completeness, the codebase also contains an optional plan-initialization prompt
that asks a model to produce a JSON object with issue nodes, element nodes, and
retrieval goals from the user query. The initialization prompt requests
1--3 missing-evidence descriptions per element and was designed for exploratory
diagnostics rather than for the final reported configuration.

Its core instruction is:

\begin{quote}\small
\texttt{Given the user query for legal consultation, identify the legal issues,}\\
\texttt{required legal elements, and retrieval goals. Output a JSON object with}\\
\texttt{three arrays: issues, elements, and retrieval\_goals. Each element must}\\
\texttt{be attached to an issue and should include 1--3 missing-evidence}\\
\texttt{descriptions. Focus on legally relevant issues only.}
\end{quote}

\subsection{Dataset Construction}

The main benchmark is a 50-case synthetic Chinese labor-law consultation pilot.
The generator draws from eight case templates:
wrongful dismissal, wage arrears, work injury, contract-breach compensation,
non-compete, overtime, social-insurance contribution disputes, and year-end
bonus disputes. Each synthetic dialogue contains an initial user query, a case
type label, a list of ground-truth legal elements, a list of key statutes, and
1--3 clarification turns. The synthetic dataset is used to test retrieval
control behavior and not to claim deployment-level legal quality.

\subsection{Retriever and Evidence Sources}

The retriever uses a simulated Chinese legal corpus containing labor statutes,
contract provisions, representative cases, and short policy-style web notes.
For the MAP agent, statute retrieval returns up to three items per action.
Case retrieval in the corpus API returns up to two items per action; the fixed
round and no-graph baselines use the same retriever family. Evidence alignment
is implemented through lightweight character-level overlap and legal-term
matching, which is why the paper treats the evidence linker as a calibrated but
still imperfect internal metric.

\subsection{Hyperparameters and Main Variants}

Unless otherwise stated, all runs use
$\theta_e=0.85$, $\theta_{ev}=0.70$, $\delta=0.05$, and a maximum round budget
$r_{\max}=7$. The rule, no-threshold-stop, no-graph, and fixed-round baselines
are deterministic offline runs over the same fixed legal-element templates.
Model-backed action-selection runs use temperature $0.3$ and the same fixed
templates. The main model-backed result uses DeepSeek V4-Pro; the cross-model
appendix compares the same configuration against Qwen-Plus, GLM-5.1, and
MiniMax-M2.7.

\subsection{Diagnostic Subsets and Extension Runs}

The appendix reports three additional experiment families beyond the main
50-case pilot.

\paragraph{Preliminary 20-case no-stop check.}
We build a 20-case diagnostic subset by round-robin sampling across case types
from the main dataset. On this subset, we compare the normal policy against a
hard-no-stop variant that is forbidden from emitting
\texttt{generate\_conclusion} before exhausting the round budget. We test
$r\in\{3,5,7\}$ for both the deterministic rule policy and Qwen-Plus.

\paragraph{Hard-case budget sweep.}
We construct the hard-case subset by selecting all dialogues whose original
rule-based run has EC $< 1.0$ on the 50-case pilot, which yields 11 cases after
sorting by case type and dialogue id. We then compare the normal policy and the
hard-no-stop policy at budgets $r\in\{1,2,3,5,7\}$. This is the main stopping
analysis reported in the paper because it has stronger resolution than the
preliminary 20-case check.

\paragraph{Multi-seed robustness.}
To test sensitivity to synthetic dataset composition, we generate four
independent 30-case pilots with seeds $\{42, 7, 21, 84\}$. On each pilot we run
the deterministic rule policy and the Qwen-Plus model-backed policy under the
same fixed-schema setting and aggregate EC, EVC, rounds, evidence count, and
token usage.

\subsection{Reproducibility Scope}

The appendix is intended to make the reported experiments auditable. The exact
scripts used to generate the hard-case subset, the stratified 20-case subset,
and the corresponding figures are included in the project release. Since the
benchmark is synthetic and the retriever is simulated, the main value of this
appendix is to clarify what was held fixed across runs and what was varied in
each diagnostic.

\clearpage
\section{Experimental Tables}

\begin{table}[H]
\centering
\caption{Detailed case-type diagnostic values corresponding to Appendix Figure~\ref{fig:appendix-case-diagnostics}.}
\label{tab:appendix-case-diagnostics}
\footnotesize
\begin{tabular}{lrrrrrr}
\toprule
Case type & $n$ & Rule EC & No-stop EC & V4-Pro EC & Rule rounds & V4-Pro rounds \\
\midrule
Contract breach & 11 & 1.000 & 1.000 & 1.000 & 4.00 & 4.00 \\
Work injury & 10 & 1.000 & 1.000 & 1.000 & 4.00 & 4.00 \\
Year-end bonus & 3 & 1.000 & 1.000 & 1.000 & 4.00 & 3.00 \\
Wage arrears & 10 & 1.000 & 1.000 & 1.000 & 4.00 & 3.00 \\
Social insurance & 8 & 0.000 & 0.000 & 1.000 & 7.00 & 2.62 \\
Non-compete & 3 & 0.000 & 0.000 & 1.000 & 7.00 & 3.00 \\
Wrongful dismissal & 5 & 1.000 & 1.000 & 1.000 & 4.00 & 3.00 \\
\bottomrule
\end{tabular}
\end{table}

\vspace{-0.5em}

\begin{table}[H]
\centering
\caption{Hard-case budget sweep over the 11 rule-failure cases from the original 50-case pilot.}
\label{tab:appendix-hardcase}
\footnotesize
\resizebox{\textwidth}{!}{
\begin{tabular}{llcccccc}
\toprule
Policy & Budget & EC & EVC & Rounds & Evidence & Tokens & Latency (s) \\
\midrule
Rule & 1 & 0.000 & 1.000 & 1.0 & 3.0 & -- & -- \\
Rule & 2 & 0.000 & 1.000 & 2.0 & 6.0 & -- & -- \\
Rule & 3 & 0.000 & 1.000 & 3.0 & 9.0 & -- & -- \\
Rule & 5 & 0.000 & 1.000 & 5.0 & 15.0 & -- & -- \\
Rule & 7 & 0.000 & 1.000 & 7.0 & 21.0 & -- & -- \\
Qwen-Plus & 1 & 0.545 & 1.000 & 1.0 & 3.0 & 569.27 & 5.20 \\
Qwen-Plus & 2 & 1.000 & 1.000 & 2.0 & 5.73 & 1311.00 & 13.83 \\
Qwen-Plus & 3 & 1.000 & 1.000 & 2.91 & 5.73 & 2241.36 & 18.57 \\
Qwen-Plus & 5 & 1.000 & 1.000 & 2.82 & 5.45 & 2201.27 & 16.09 \\
Qwen-Plus & 7 & 1.000 & 1.000 & 2.82 & 5.45 & 2220.09 & 17.77 \\
Qwen-Plus hard-no-stop & 1 & 0.591 & 1.000 & 1.0 & 3.0 & 556.27 & 4.46 \\
Qwen-Plus hard-no-stop & 2 & 1.000 & 1.000 & 2.0 & 6.0 & 1295.18 & 10.92 \\
Qwen-Plus hard-no-stop & 3 & 1.000 & 1.000 & 3.0 & 9.0 & 2389.27 & 23.07 \\
Qwen-Plus hard-no-stop & 5 & 1.000 & 1.000 & 5.0 & 15.0 & 5006.18 & 29.72 \\
Qwen-Plus hard-no-stop & 7 & 1.000 & 1.000 & 7.0 & 21.0 & 8460.64 & 55.74 \\
\bottomrule
\end{tabular}
}
\end{table}

\vspace{-0.5em}

\begin{table}[H]
\centering
\caption{Multi-seed robustness over independently generated 30-case synthetic pilots.}
\label{tab:appendix-seed}
\footnotesize
\begin{tabular}{lcccccc}
\toprule
Policy & Seed & EC & EVC & Rounds & Evidence & Tokens \\
\midrule
Rule & 42 & 0.700 & 1.000 & 4.9 & 12.6 & -- \\
Rule & 7 & 0.667 & 1.000 & 5.0 & 13.0 & -- \\
Rule & 21 & 0.733 & 1.000 & 4.8 & 12.2 & -- \\
Rule & 84 & 0.533 & 1.000 & 5.4 & 14.6 & -- \\
Qwen-Plus & 42 & 1.000 & 1.000 & 3.30 & 6.9 & 2939.23 \\
Qwen-Plus & 7 & 1.000 & 1.000 & 3.17 & 6.5 & 2759.00 \\
Qwen-Plus & 21 & 1.000 & 1.000 & 3.30 & 6.9 & 2925.63 \\
Qwen-Plus & 84 & 1.000 & 1.000 & 2.97 & 5.9 & 2518.17 \\
\bottomrule
\end{tabular}
\end{table}

\FloatBarrier
\clearpage

\section{Supplementary Figures}
\vspace{-0.6em}

\begin{center}
\begin{minipage}{0.46\textwidth}
\centering
\includegraphics[width=\linewidth]{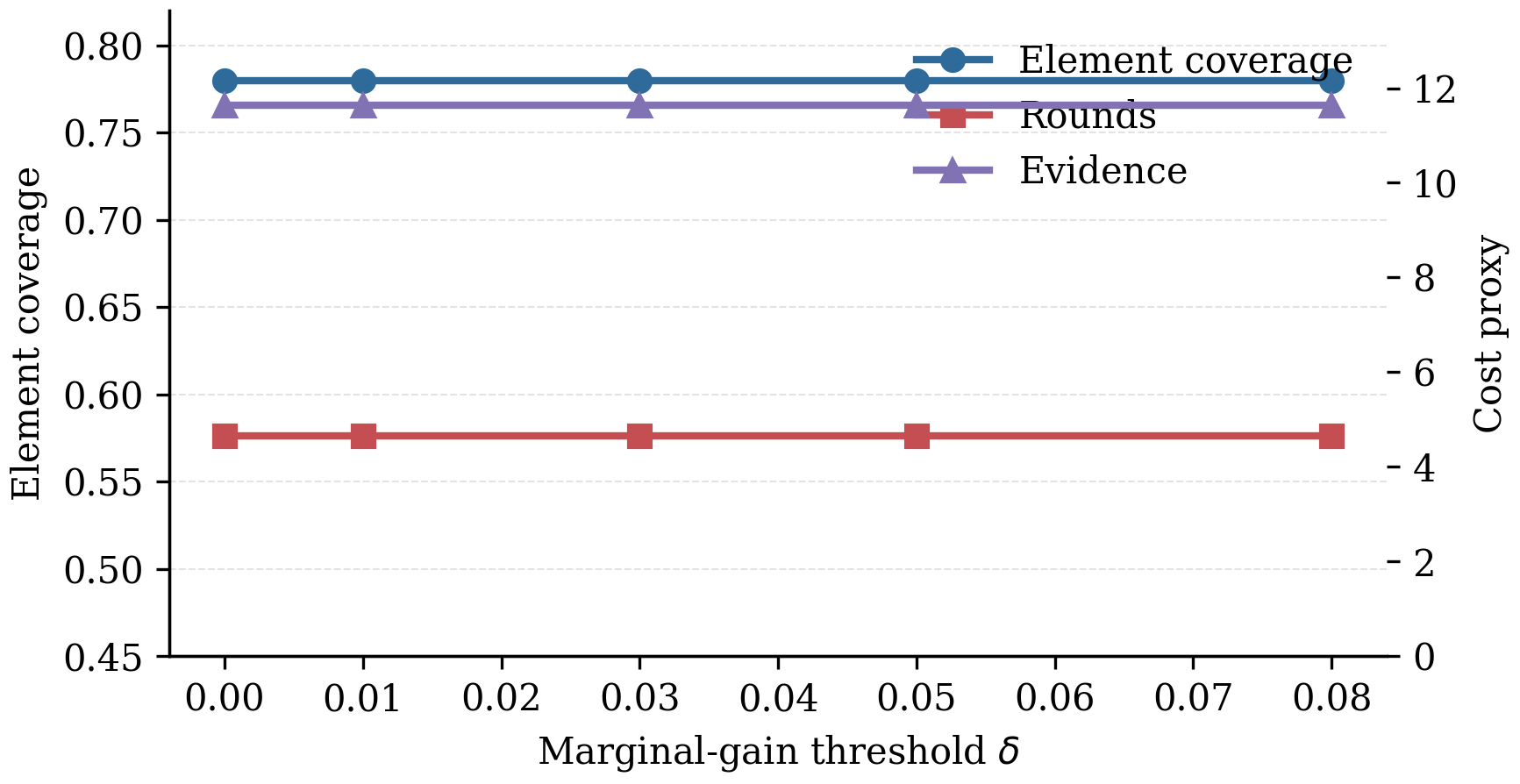}\\[-0.4em]
{\small (a) Threshold sensitivity}
\end{minipage}
\hfill
\begin{minipage}{0.46\textwidth}
\centering
\includegraphics[width=\linewidth]{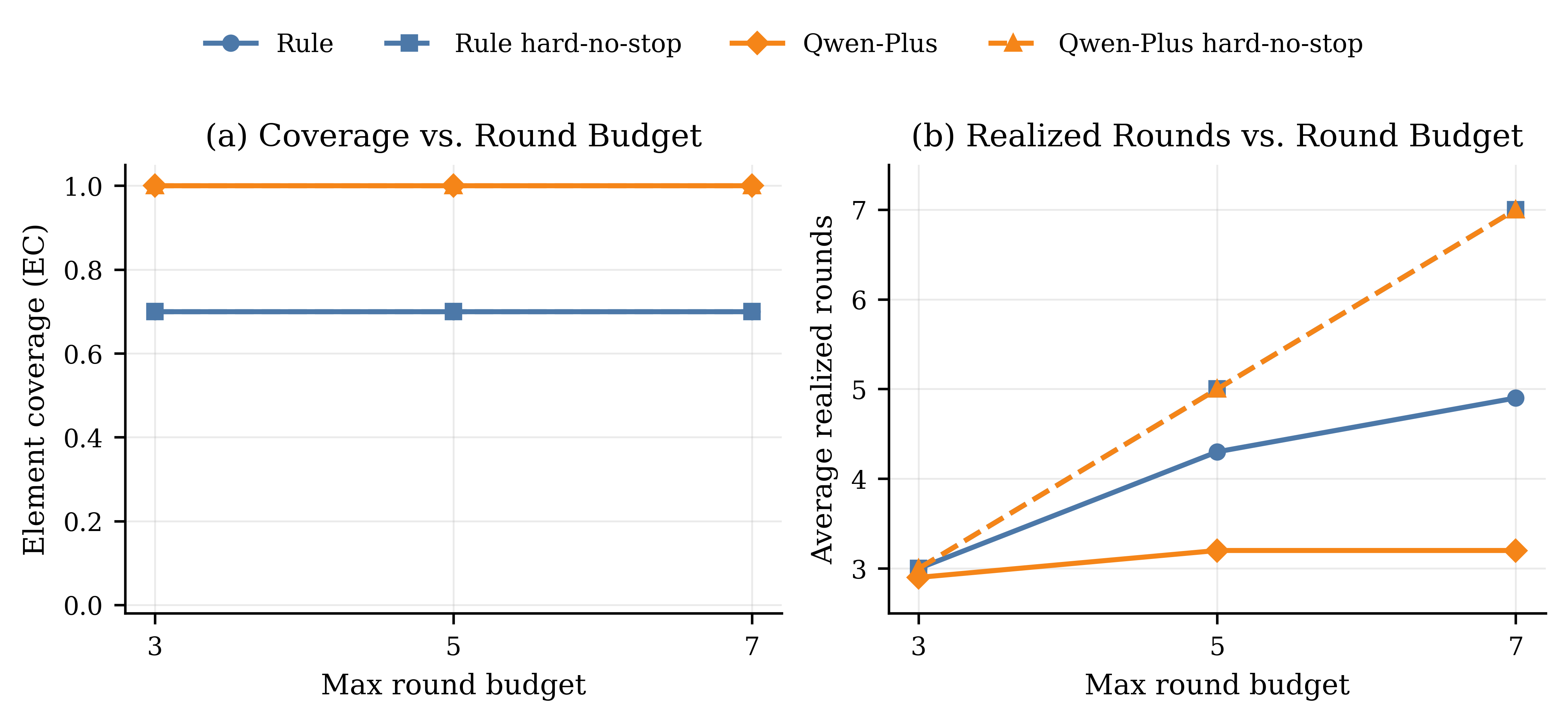}\\[-0.4em]
{\small (b) Preliminary no-stop check}
\end{minipage}

\vspace{0.3em}

\begin{minipage}{0.58\textwidth}
\centering
\includegraphics[width=\linewidth]{fig_seed_robustness.png}\\[-0.4em]
{\small (c) Multi-seed robustness}
\end{minipage}

\vspace{-0.2em}
\captionof{figure}{Supplementary diagnostics. 
(a) Threshold sensitivity of the rule-only policy after Chinese character-level marginal-gain scoring.
(b) Preliminary 20-case no-stop consistency check, where coverage is flat and the main effect is additional retrieval cost.
(c) Multi-seed robustness over independently generated synthetic pilots.}
\label{fig:appendix-supplementary-diagnostics}
\end{center}

\vspace{-0.4em}

\begin{center}
\includegraphics[width=0.72\textwidth]{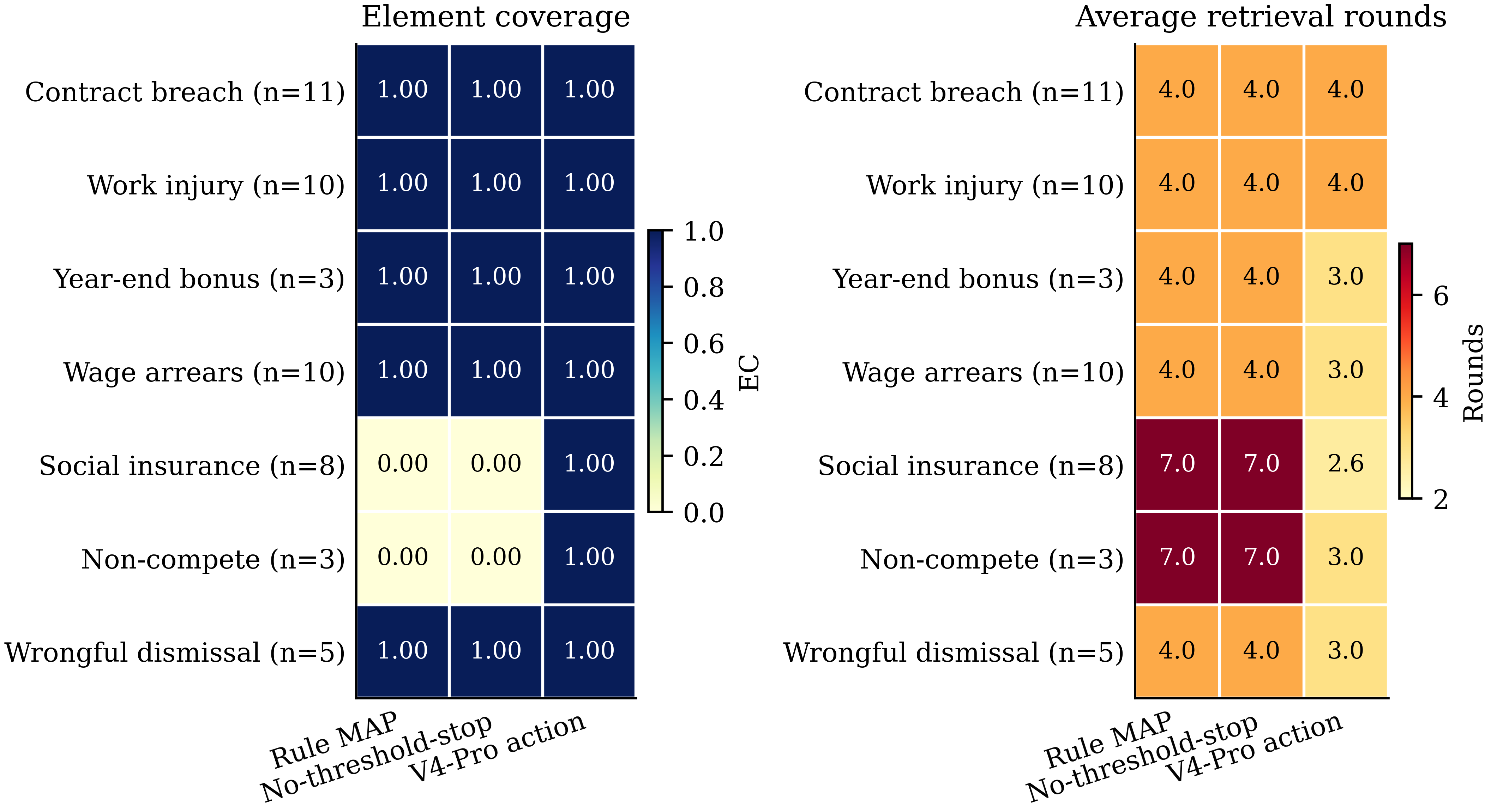}
\vspace{-0.4em}
\captionof{figure}{Case-type diagnostic matrix under the fixed-schema synthetic pilot.
The left panel reports measured element coverage by case type and system; the
right panel reports average retrieval rounds. Deterministic control covers
routine categories but fails on social-insurance and non-compete disputes,
suggesting that these failures are driven by action-selection or
evidence-alignment limitations rather than premature threshold stopping.
V4-Pro action selection recovers these categories with fewer rounds under the
same fixed legal-element initialization.}
\label{fig:appendix-case-diagnostics}
\end{center}

\FloatBarrier

\clearpage
\bibliography{references}
\bibliographystyle{plainnat}

\end{document}